\documentclass[sigconf, nonacm]{acmart}
\AtBeginDocument{%
  \providecommand\BibTeX{{%
    Bib\TeX}}}

\def\BibTeX{{\rm B\kern-.05em{\sc i\kern-.025em b}\kern-.08em
    T\kern-.1667em\lower.7ex\hbox{E}\kern-.125emX}}

\usepackage{amsmath,amsfonts}
\usepackage{algorithmic}
\usepackage{graphicx}
\usepackage{textcomp}
\usepackage{hyperref}       
\usepackage{cleveref}
\usepackage{url}            
\usepackage{booktabs}       
\usepackage{nicefrac}       
\usepackage{microtype}      
\usepackage{multirow}
\usepackage{tabularx} 
\usepackage{makecell} 
\usepackage{comment}
\usepackage{caption}
\usepackage{subcaption}
\usepackage{enumitem}
\usepackage[table,xcdraw]{xcolor}
\usepackage{soul}
\usepackage{fontawesome5} 
\begin{document}

\title{HARMONI: Multimodal Personalization of Multi-User Human-Robot Interactions with LLMs}


\author{Jeanne Malécot$^{*,~1,~2}$, Hamed Rahimi$^{*,~2}$, Jeanne Cattoni$^{3}$, Marie Samson$^{2}$, Mouad Abrini$^{2}$, Mahdi Khoramshahi$^{2}$, Maribel Pino$^{3}$, Mohamed Chetouani$^{2}$}
 \authornote{Both first authors contributed equally to this research. Jeanne Malecot is currently with Institut Curie, and this work was carried out during her internship at ISIR under the supervision of Hamed Rahimi and Mohamed Chetouani. \\Correspondence: rahimi@isir.upmc.fr}
 \affiliation{%
   \institution{$^1$Institut Curie, Université Paris-Saclay\\$^2$Institute of Intelligent Systems and Robotics (ISIR), Sorbonne University\\$^3$Assistance Publique – Hôpitaux de Paris (AP-HP), Université Paris Cité}
   \city{Paris}
   \country{France}
 }








\renewcommand{\shortauthors}{Malécot and Rahimi et al.}

\begin{abstract}
Existing human-robot interaction systems often lack mechanisms for sustained personalization and dynamic adaptation in multi-user environments, limiting their effectiveness in real-world deployments. We present HARMONI, a multimodal personalization framework that leverages large language models to enable socially assistive robots to manage long-term multi-user interactions. The framework integrates four key modules: (i) a perception module that identifies active speakers and extracts multimodal input; (ii) a world modeling module that maintains representations of the environment and short-term conversational context; (iii) a user modeling module that updates long-term speaker-specific profiles; and (iv) a generation module that produces contextually grounded and ethically informed responses. Through extensive evaluation and ablation studies on four datasets, as well as a real-world scenario-driven user-study in a nursing home environment, we demonstrate that HARMONI supports robust speaker identification, online memory updating, and ethically aligned personalization, outperforming baseline LLM-driven approaches in user modeling accuracy, personalization quality, and user satisfaction.
\end{abstract}

\maketitle

\faGithub  ~~ \href{https://github.com/hamedR96/HARMONI/}{Code and Data}

\section{Introduction}

\begin{figure}[t]
    \centering
    \includegraphics[width=1\linewidth]{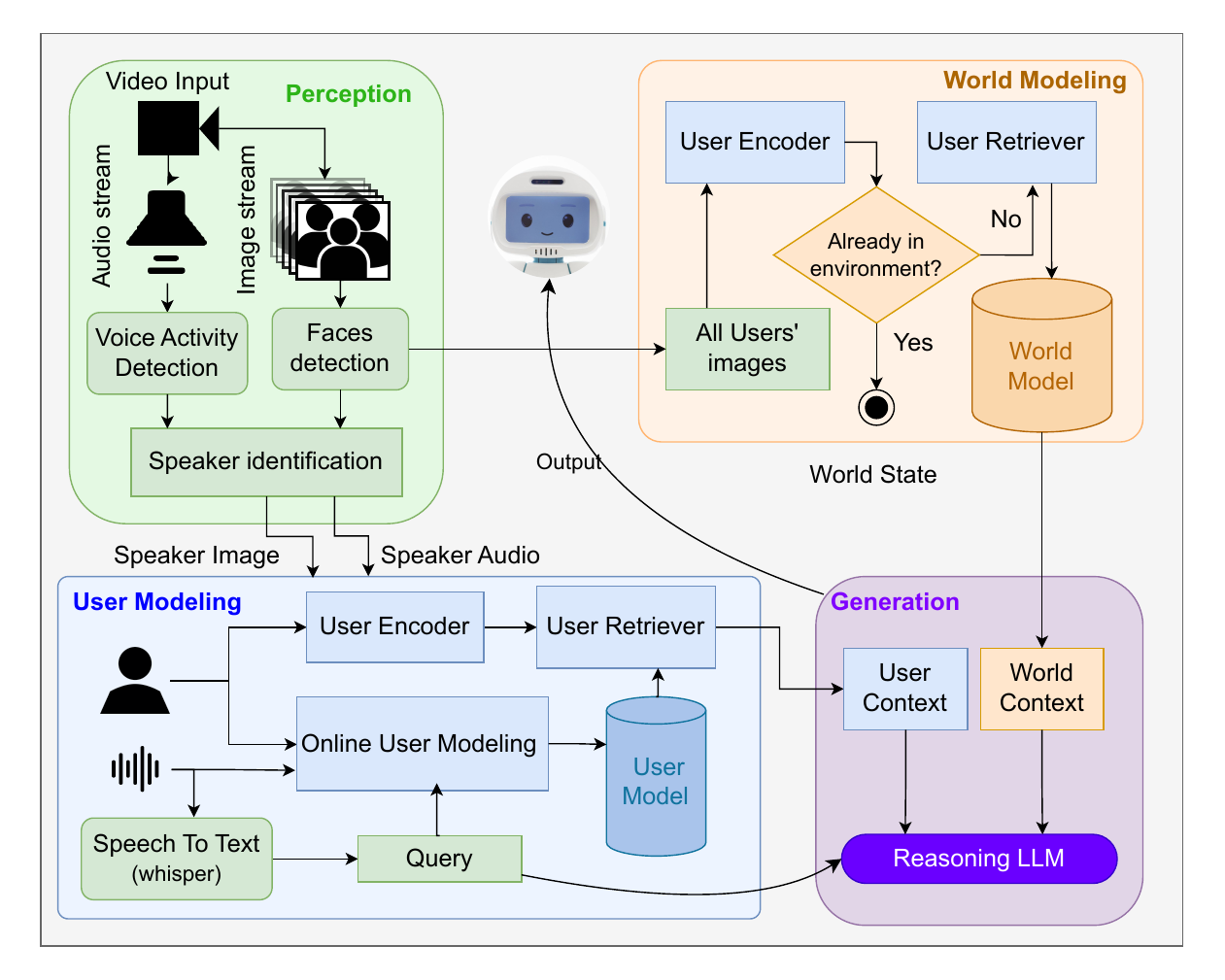}
    \caption{\textbf{Overview of HARMONI}. 
    The proposed framework consists of four modules:
    (i) a \textit{multimodal perception module} that identifies the active speaker and extracts the query from visual and auditory inputs; (ii) a \textit{world modeling module} that maintains representations of all users and the ongoing conversation within a session; (iii) a \textit{user modeling module} that retrieves and updates speaker-specific profiles for long-term personalization; and (iv) a \textit{generation module} that produces contextually grounded and personalized responses conditioned on the speaker, their profile, surrounding users, and the conversational context.}
    \label{fig:arch}
\end{figure}

Socially Assistive Robotics (SAR)~\cite{feil2011socially} has emerged as a promising field aimed at supporting individuals facing diverse challenges- including older adults, patients, and people with disabilities- by addressing their physical, cognitive, and social needs. While much of the research in this area has focused on one-on-one Human-Robot Interactions (HRI)~\cite{rahimi2025user_1,rahimi2025user_2}, there is a growing demand for robots that can function effectively in multi-user environments~\cite{louie2014autonomous,bettencourt2025investigating}, such as nursing homes, therapy centers, and group rehabilitation programs. 
%
%
\medbreak

\begin{figure*}[h]
    \centering
    \includegraphics[width=0.8\linewidth]{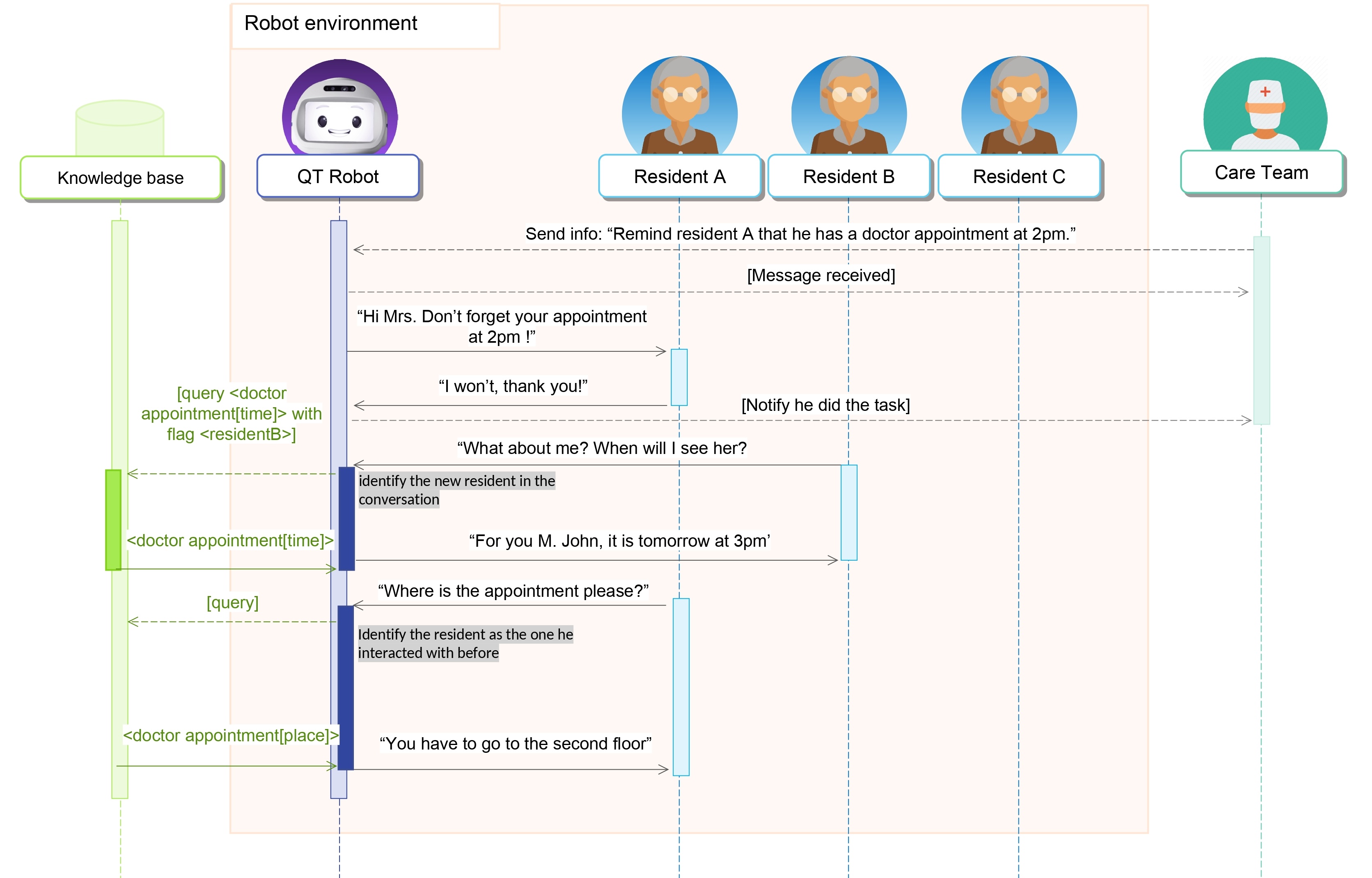}
    \caption{\textbf{Overview of a challenging multi-turn interaction scenario.} The robot dynamically switches context to a new user (Resident B) upon interruption and accurately resolves anaphora (e.g., ``the appointment'') in follow-up queries by maintaining the updated conversational state, clearly distinguishing it from the initial interaction with Resident A.}

    \label{fig:scenario}
\end{figure*}

Existing HRI systems, however, often lack robust mechanisms for personalization across multiple turns and multiple users~\cite{li2021dialogue,9515550}. 
A robot must coordinate dynamically, adapt to evolving environments, and adjust behaviors in response to multiple interacting users~\cite{umbrico2022design}. Without effective personalization, these systems risk reduced engagement, diminished trust, and limited long-term effectiveness~\cite{lee2012personalization}. Multi-user adaptation over time including memory of prior interactions, recognition of user states, and flexible reuse of past experiences is critical to maintaining user interest, building rapport, and ensuring the acceptability of SARs \cite{finkel2023robot}. To this day, however, such robots remain limited in their ability to provide deep, long-term personalization across multiple users and interaction sessions, highlighting a critical gap in current HRI systems.

\medbreak

To address these limitations, we propose a multimodal personalization framework for multi-user HRI that enables robots to dynamically adapt to both individual and group needs, while ensuring compliance with ethical and social constraints. As illustrated in \Cref{fig:arch}, the framework first receives a video-based query from the user and processes it through modular components operating in parallel to generate personalized dialogue responses. These responses leverage both short-term conversational context and long-term adaptation, enabling robots to recall past interactions and dynamically generate future behaviors through user modeling. This work advances the development of SARs by introducing a framework that integrates personalization with privacy- and ethics-preserving mechanisms, enabling deployment in sensitive contexts such as elder care. The key contributions of this paper are: \textbf{(1)} a multimodal personalization strategy that adapts to diverse user preferences and behaviors; \textbf{(2)} an online user modeling approach for adaptive and continuous personalization; \textbf{(3)} a real-time multi-user memory management system encompassing both long-term memory, which supports sustained adaptation across sessions, and short-term memory, which captures context within the present interaction; \textbf{(4)} an ethical design methodology that explicitly balances personalization with privacy and safety considerations; and \textbf{(5)} a scenario-driven evaluation conducted in realistic nursing home environments, \Cref{fig:scenario}, demonstrating the effectiveness and challenges of SARs in multi-user, socially assistive settings.

\section{Related Work}
\subsection{Personalization in Dialogue Systems}
Personalization in dialogue systems has been studied extensively to enhance user engagement and response quality. Prior work spans several dimensions: (i) surveys of datasets and problem categories~\cite{chen2024recent}, (ii) fine-tuning frameworks such as In-Dialogue Learning (IDL) that infer personas from dialogue history without predefined profiles~\cite{cheng2024dialogues}, (iii) latent-variable models like MIRACLE that decompose complex traits into multi-faceted attributes~\cite{lu2023miracle}, (iv) attention-based mechanisms such as Persona-Adaptive Attention (PAA) that jointly model persona and context while reducing redundancy~\cite{huang2022personalized}, and (v) prompt-tuning approaches that enable efficient, multilingual personalization in large-scale pre-trained models~\cite{kasahara2022building}. While these advances substantially improve single-user, text-based personalization, challenges remain in multi-user, socially assistive settings where privacy, ethical safety, and long-term memory are critical.

\subsection{Multi-User Human-Robot Interactions}
Multi-user HRI research addresses challenges such as managing interruptions, coordinating overlapping tasks, and supporting collaborative decision-making. Approaches include intent-based interruption detection with LLMs and context-aware strategies~\cite{jung2021interruption}, multi-user dialogue modeling via contextual query rewriting on datasets like Multi-User MultiWOZ~\cite{chen2023multiuser}, and analyses of interruption effects showing that extrinsic disruptions increase perceived workload despite stable task performance~\cite{dahiya2023impact}. Collaborative task execution frameworks, such as activation spreading architectures, further enable dynamic task allocation across overlapping sub-tasks~\cite{anima2019collaborative}. While these works advance interruption management, decision-making, and collaboration, they emphasize coordination over personalization. In contrast, our framework introduces \emph{multimodal personalization} and \emph{real-time multi-user memory management}, allowing robots to adapt to individual user profiles and conversational histories in addition to coordinating across users.

\subsection{Multimodal Memory Management}
Memory architectures and multimodal feature extraction are critical for maintaining coherence in adaptive dialogue systems. Recent work includes dynamic frameworks such as User-VLM R1~\cite{rahimi2025reasoning} and Mem0, which consolidate and retrieve salient information with low latency~\cite{chhikara2024mem0}, and graph-based approaches that improve relational reasoning through entity–relation modeling. In multimodal dialogue, position-aware fusion models like PMATE integrate textual and visual signals for improved response grounding~\cite{he2024multimodal}, while embodied agents benefit from context- and environment-aware planning frameworks such as CAPEAM~\cite{kim2023context}. Complementary studies further show that incorporating visual cues (e.g., facial expressions, gaze)~\cite{rahimi2025demographic} with textual inputs enhances intent disambiguation and emotion recognition~\cite{multimodal2023learning}. Although these advances underscore the value of memory and multimodal integration, they are largely restricted to single-user or task-specific settings. Our framework extends this direction by addressing \emph{multi-user socially assistive HRI}, combining multimodal personalization, dynamic user modeling across temporal scales, and ethically aligned memory management for sensitive domains such as elder care.

\subsection{Ethical Challenges in SAR}
Ethical considerations are critical for SARs and conversational AI, especially in healthcare, education, and eldercare. Foundational principles—autonomy, beneficence, non-maleficence, and justice- highlight risks related to deception, consent, and substitution of human care~\cite{feil2011ethical}. Studies in diverse and under-resourced contexts reveal challenges including language barriers, unintended harms, trust issues, and feasibility constraints~\cite{sharma2022ethical}. Privacy and data protection are paramount, motivating privacy-by-design practices such as encryption, data minimization, explicit consent, and transparent governance~\cite{privacy2024dark}. Addressing bias, inclusivity, and user trust requires diverse datasets, inclusive design, and transparency in anthropomorphic interactions~\cite{social2024ethical}. Our framework builds on these insights by embedding privacy- and ethics-preserving mechanisms into a multimodal personalization pipeline, enabling SARs to adapt safely and transparently to both individual and group needs.

\medbreak

We note that prior works in the literature typically address these issues in a disjoint manner, focusing on each aspect separately. In contrast, in this work, we propose a unified approach that simultaneously addresses all of these requirements. Our work addresses a gap in current research by integrating multimodal personalization, real-time multi-user memory management, and ethically aligned design for socially assistive human-robot interactions. While existing dialogue systems focus on single-user interactions and multi-user HRI research emphasizes task coordination and interruption management, our framework enables robots to adapt to both individual preferences and collective dynamics. By combining adaptive interactions with privacy, ethical safety, and long-term engagement, our approach supports responsible deployment in sensitive contexts such as elder care.

\begin{figure*}
    \centering
    \includegraphics[width=\linewidth]{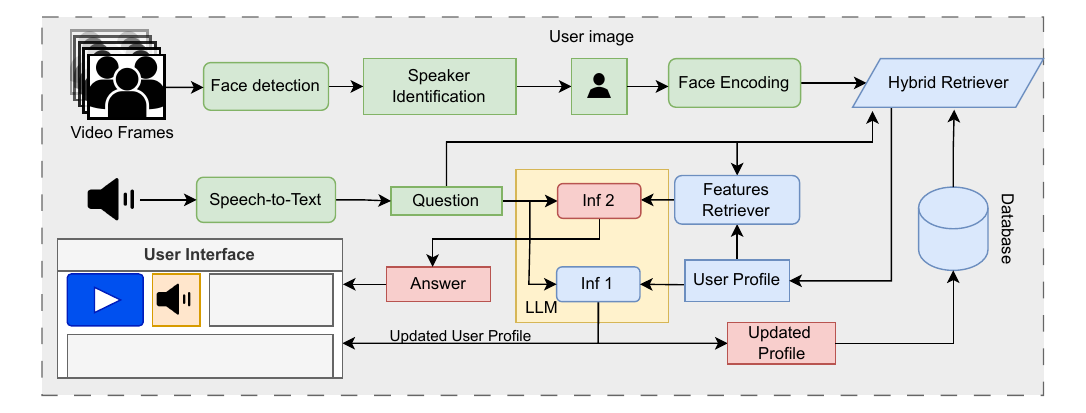}
    \caption{\textbf{Overview of the proposed user interface architecture.} The interface integrates (i) \textit{Perception} (green components), which identifies active speakers and transcribes input; (ii) \textit{User Modeling} (blue components), which retrieves and updates user profiles with long-term memory; (iii) and \textit{Generation} (red components), which produces contextually grounded and personalized responses. The \textit{World Modeling}, which maintains short-term conversational context and environmental state, is stored at the user interface for simplicity. The interface transparently displays updated user profiles and dialogue history, supporting explainability and real-time adaptation in multi-user scenarios.}
    \label{fig:interface}
\end{figure*}

\section{Methods}
To address the challenges of multi-turn, personalized dialogue in shared human-robot environments, we design a scenario-driven modular system that integrates perception, user modeling, world modeling, and response generation. Our approach operates in nursing home scenarios where multiple residents and staff interact simultaneously, creating overlapping conversational signals and heterogeneous objectives.

\subsection{Scenario}
We designed custom scenarios to simulate the behavior of our system in the context of nursing homes. These scenarios capture different types of exchanges that could occur in a shared room, involving at least one resident. The main challenges highlighted are the system’s ability to recall the appropriate information about each user depending on the context, and to manage multiple users simultaneously, maintaining distinct conversation threads without confusing user profiles. 
As shown in \Cref{fig:scenario}, we consider a challenging nursing home scenario where a socially assistive robot must manage dynamic, multi-turn dialogues while performing task-oriented actions. The interaction begins with a directed reminder for a user, the Patient A, regarding an upcoming appointment, establishing a straightforward one-to-one dialogue. The complexity arises when another user, the Patient B, initially uninvolved, interjects with a self-referential query, requiring the robot to detect the speaker change and update its conversational context in real-time. The system must correctly resolve deictic references and anaphoric expressions, linking utterances such as ``Where is the appointment?'' to the relevant antecedent (the Patient B's appointment) rather than the original context. This scenario highlights the model's capability to perform on-the-fly context switching, dialogue grounding, and knowledge base querying, thereby demonstrating robust handling of spontaneous interruptions and multi-turn interactions, which are critical for reliable operation in real-world, multi-user human-robot environments.

\subsection{Problem Formulation}

Let $\mathcal{U}$ denote the dynamic set of all users in the system, where new users $u_k$ ($k=1,\dots,n$) can be created and added to $\mathcal{U}$. Interactions unfold in discrete time steps $t = 1,2,\dots,T$. At each step $t$, an active user $u_t \in \mathcal{U}_t \subseteq \mathcal{U}$ issues a natural-language query $q_t$ to the assistive robot $a$, where $\mathcal{U}_t$ denotes the set of users that are present or interacting with the system at time $t$.

The robot maintains two internal states:  
(i) a \textit{world state} $w_t \in \mathcal{W}$, which captures the current environment, including the users present, as well as a short-term dialogue memory; and  
(ii) a \textit{user model} $p_{u,t} \in \mathcal{P}$, which encodes user-specific information, such as a profile and a long-term memory of past conversations. Both models are dynamically updated based on the robot’s observations and user inputs.  

Given a query $q_t$ from user $u_t$, the robot forms an observation
\begin{equation}
    o_{u,t} \triangleq \big(q_t,\; p_{u_t,t},\; w_t\big) \in \mathcal{O}.
\end{equation}

Based on this observation, the robot generates a response $r_t \in \mathcal{R}$ according to a policy parameterized by an LLM:
\begin{equation}
    r_t \sim \pi_\theta\!\left(\cdot \,\middle|\, o_{u,t}\right),
\end{equation}
where $\pi_\theta$ produces the response token-by-token.  

In parallel, the robot performs a \textit{system update}. First, the user model is refined by incorporating information contained in the query, according to a structured update policy $\pi_\phi$:
\begin{equation}
    p_{u,t+1} \sim \pi_\phi\!\left(\cdot \,\middle|\, p_{u,t}, q_t\right).
\end{equation}
Second, the conversation history of $u_t$ is extended by appending the latest interaction turn.


\subsection{System Design}
As illustrated in \Cref{fig:arch}, the proposed system design consists of four primary modules. The (i) \textit{Perception Module} is responsible for interpreting and analyzing video input, including speaker detection. The (ii) \textit{User Modeling Module} manages user-related information by creating or retrieving a user profile, maintaining long-term interaction history, and updating user-specific data for future interactions. The (iii) \textit{World Modeling Module} focuses on interpreting the situational and environmental context exhibited by the robotic agent, while also maintaining short-term memory to capture recent events. Finally, the (iv) \textit{Generation Module} produces privacy-aware responses that leverage both short-term and long-term memory, ensuring coherent and contextually appropriate interactions.
  
\medbreak
\subsubsection{\textbf{Perception Module}}
The primary task of this module is to interpret video input by extracting the spoken question and identifying the active speaker. This process is carried out by synthesizing the video stream into both audio and image sequences with Voice Activity Detection (\textit{FastRTC}~\cite{boulton2025fastrtc}). The video stream is processed to identify potential users by extracting facial regions with a face detection model (\textit{YOLOv8-Face-Detection}~\cite{yusepp2025yolov8face}). The active speaker is then determined by correlating facial movements, particularly lip motion (\textit{shape-predictor-68-face-landmarks}). 

\medbreak
\subsubsection{\textbf{User Modeling Module}}

This module is responsible for retrieving and maintaining user-specific profiles and long-term memory. Given the identified speaker’s image and corresponding audio, the system first transcribes the spoken question using a speech-to-text generative model (\textit{~Whisper-Turbo}~\cite{radford2023robust}). In parallel, the speaker’s image is encoded into a vector representation using the INSIGHTFACE model (\textit{buffalo\_l}~\cite{ren2023pbidr}) and compared against entries in the user database. If the user does not exist in the database, a new profile is initialized; otherwise, the system retrieves the corresponding features of the user profile along with relevant memory of past interactions based on semantic similarity to the current question using a text encoding model \textit{(google/EmbeddingGemma-300m)~\cite{choi2025embeddinggemma}}. Additionally, the user image, transcribed question, and audio are passed to a feature extraction component that estimates demographic and affective attributes such as age, gender, and emotion. These features are used to initialize or dynamically update the user profile throughout the course of interactions. The integration and reasoning over these multimodal features are supported by a large language model (\textit{google/gemma3-27b}~\cite{team2025gemma}).

\medbreak
\subsubsection{\textbf{World Modeling Module}} 
This module is designed to maintain short-term conversational sessions and to retrieve the most relevant memories associated with user profiles present in the environment. The input to this module consists of all users detected in the video stream. For each detected user, the module retrieves the corresponding segment of the world model by leveraging text-based similarity. Specifically, it employs the same text embeddings used in the User Modeling Module to compute encoding similarity in the language modality, ensuring consistent and semantically aligned retrieval. This allows HARMONI to reduce hallucinations and explicitly indicate uncertainty through its modular design, which separately encodes user profile and the joint human–robot context, thereby enabling more reliable reasoning in interactive tasks~\cite{kalai2025language}.

\medbreak

\subsubsection{\textbf{Generation Module}}
This module receives as input the current observation, which includes the user’s question, retrieved user memory, relevant profile features, and the world model containing short-term conversational history and contextual information from the environment. Based on these inputs, a large language model generates a response that is conditioned on both the user profile and the associated memory, ensuring contextually appropriate and personalized outputs. In addition to response generation, the module incorporates mechanisms for \emph{ethics and privacy preservation}. Specifically, it filters sensitive or personally identifiable information before inclusion in responses, enforces safety constraints to avoid harmful or biased outputs, and adheres to user-defined privacy preferences. These safeguards ensure that generated responses are not only accurate and contextually aligned but also responsible and privacy-aware.

\subsection{\textbf{User Interface Implementation}}
As shown in \Cref{fig:interface}, the proposed system is deployed with a user interface that supports multi-user interactions with the robot. The robot is capable of recognizing speakers, encoding their facial features, retrieving their profiles, and selecting relevant attributes to generate personalized responses. Long-term memory is maintained within the user profiles, while short-term memory is stored in the interface history queue. For language generation, we utilize the \textit{Gemma3-27B}~\cite{team2025gemma} model via the \textit{Ollama} LLM provider~\cite{marcondes2025using} (additional base models are explored in the next section). During inference, the system first retrieves user-relevant features based on semantic similarity to the query and generates an initial response (\textit{Inf 2}). In parallel, a second inference step updates the active user’s profile with newly extracted facts from their statements and questions (\textit{Inf 1}). The final response is then presented through the user interface, which also displays the updated user profile to enhance explainability and transparency.

\section{Experiments}
To validate the performance of our proposed framework and to provide a comparison with both open-source and closed-source LLMs from multiple perspectives, including personalization quality, generation performance, scalability, and latency, we investigate the following research questions:

\begin{itemize}
    \item[\textcolor{red}{\textbf{Q1}}] \textit{How does the framework perform in user detection and user profile retrieval?}
    \item[\textcolor{red}{\textbf{Q2}}] \textit{How does the framework manage memory in user modeling and feature extraction from speaking users?}
    \item[\textcolor{red}{\textbf{Q3}}] \textit{Does the framework outperform prompt-engineered plain LLMs augmented with user models?}
    \item[\textcolor{red}{\textbf{Q4}}] \textit{What is the latency of the framework when integrated with open-source and closed-source LLMs?}
     \item[\textcolor{red}{\textbf{Q5}}] \textit{How well does the proposed framework perform in real-world and nursing home environments, and what do elderly patients think of this system?}

\end{itemize}

\subsection{Dataset}
\label{sec:dataset}
To answer \textcolor{red}{\textbf{Q1}}, we constructed a customized dataset consisting of 45 video recordings. Each video contains one to four users of different ages, with each user asking a single question. This dataset enables us to evaluate the framework's ability to (i) correctly detect the active speaker, (ii) switch seamlessly between multiple users, and (iii) retrieve the appropriate user profile. While the dataset includes six unique users in total, each video presents a different combination of these users.

To address \textcolor{red}{\textbf{Q2}}, we employed the LoCoMo dataset \cite{locomo}, leveraging the provided annotations as ground truth for evaluating the accuracy of feature extraction and user profile updates performed by the framework. The evaluation considers three aspects: (i) the number of observations made by our system compared to the dataset, (ii) the similarity of the features extracted at a specific turn, and (iii) the consistency of the global user profile over an entire session. We compare feature extraction under different conditions: updating with only the last user turn, updating with the full current session as a Short-Term-Memory (STM), and updating while retaining previous observations at each new turn as a Long-Term-Memory (LTM). 

To evaluate \textcolor{red}{\textbf{Q3}}, we utilized the PersonaFeedback dataset \cite{personafeedback}. Since the dataset was originally in Chinese, we translated it into English and subsequently into French using OPUS translation models to ensure multilingual evaluation. To integrate it into our framework, we converted the provided user profiles into the format of our database and generated answers using our personalization pipeline. 
%
%
Furthermore, to address \textcolor{red}{\textbf{Q4}}, we measured system latency and response time throughout all experiments. Finally, for \textcolor{red}{\textbf{Q5}}, the framework was deployed in a hospital setting, where standardized questionnaires were administered. Patients and experts evaluated the system in terms of performance, latency, usability, and overall satisfaction, thereby providing insights into its real-world applicability in healthcare environments.

\begin{figure*}[t]
    \centering
    \begin{subfigure}[c]{0.20\textwidth}
        \centering
        \includegraphics[width=\textwidth]{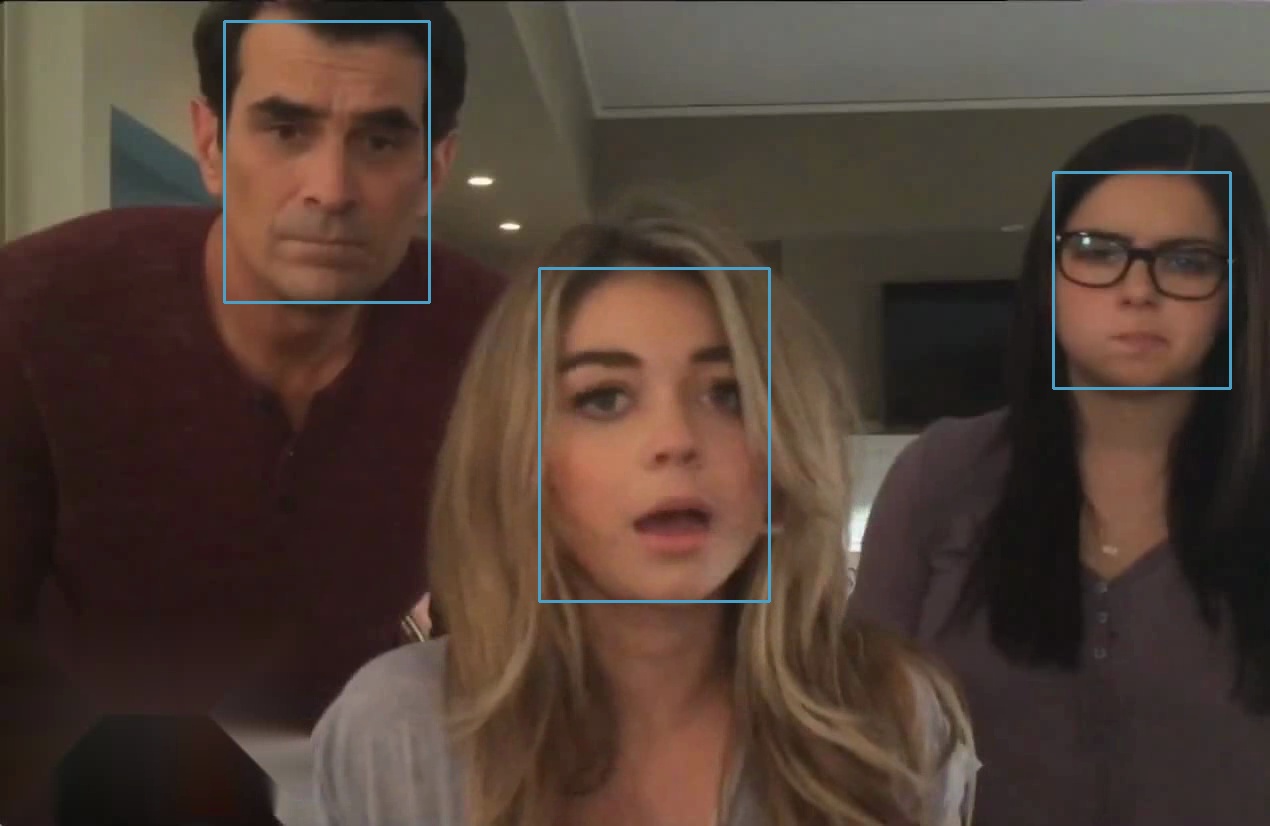}
        \caption{}
        \label{fig:test_video}
    \end{subfigure}%
    \begin{subfigure}[c]{0.80\textwidth}
        \centering
        \includegraphics[width=\textwidth]{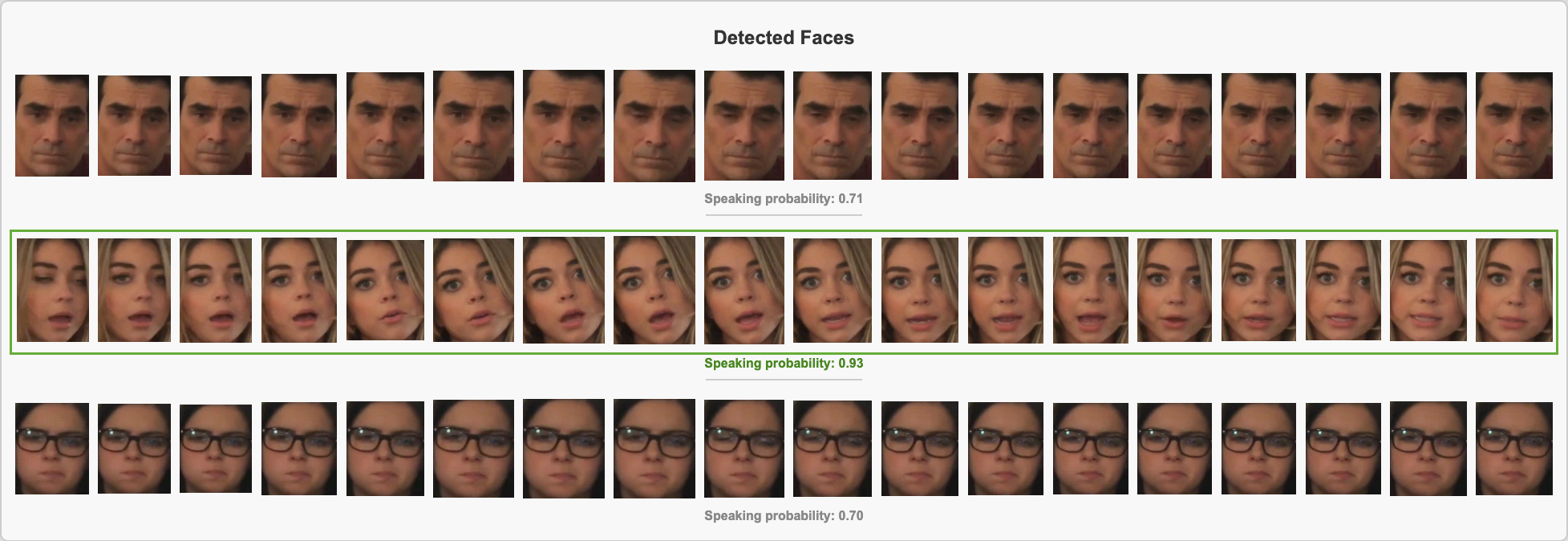}
        \caption{}
        \label{fig:face-grid}
    \end{subfigure}
    \caption{\textbf{Example of speaker detection within an ego-centric video.} 
    (a) Input video frame from egocentric perspective. 
    (b) Perception module output with face grid indicating detected speakers.}
    \label{fig:exmp}
\end{figure*}

\subsection{Metrics}
To comprehensively evaluate our framework, we employ different sets of metrics depending on the research question. For face detection, user–speaker detection, and profile retrieval (\textcolor{red}{\textbf{Q1}}), we report standard classification metrics, including F1-score, Recall, Precision, and Accuracy. For evaluating feature detection (\textcolor{red}{\textbf{Q2}}) and personalization (\textcolor{red}{\textbf{Q3}}), the primary evaluation metric is \emph{LLM-as-Judge}, which LLMs to approximate human-like evaluation when comparing different baselines. In addition, we report ROUGE and Session Similarity to broaden the evaluation scope and capture linguistic quality. We also measure inference time across all stages and compare it against baseline systems to provide a transparent view of latency and its impact on both open-source and closed-source LLMs before and after applying our framework (\textcolor{red}{\textbf{Q4}}). For the human evaluation (\textcolor{red}{\textbf{Q5}}), participants completed the System Usability Scale (SUS) questionnaire~\cite{brooke195sus,bangor2008evaluationsus} after interacting with the system. SUS is a well-established 10-item, 5-point Likert scale designed to assess perceived usability, encompassing both ease of use and overall acceptability \cite{Bevan1991WhatIU}.Beyond its straightforward administration, SUS has been shown to yield reliable results even with relatively small participant groups. Resulting scores fall within a 25–100 range, where values above 72 are generally interpreted as indicating acceptable usability. Furthermore, to assess the quality of the system’s responses, two gerontology experts rated the overall interaction on a 5-point Likert scale, ranging from ``Not satisfying at all’’ to ``Very satisfying.’’ Gerontology experts are defined as professionals who study aging and support the health, well-being, and quality of life of older adults through research, education, and practice.

\begin{table}[t]
\footnotesize	
\centering
\caption{\textbf{Evaluation of user detection and user profile retrieval over the dataset to answer (\textcolor{red}{\textbf{Q1}}).} Face detection and active speaker identification exceeds 90\% accuracy, while user retrieval  80\%. All tasks maintain low latency, with processing times under 8ms}
\begin{tabular}{lccccc}
\toprule
\textbf{Task}                & \textbf{Accuracy} & \textbf{Precision} & \textbf{Recall} & \textbf{F1} & \textbf{T(ms)} \\ \midrule \rowcolor{green!30}
Faces Detection               & 94.90             & 97.90              & 96.90           & 97.40             & 2.8            \\ \rowcolor{green!30}
Speaker Recognition  & 89.60             & 88.90              & 88.90           & 88.90             & 1.9            \\ \rowcolor{green!30}
User Retrieval & 97.90             & 93.60              & 93.60           & 93.60             & 3.12            \\ \bottomrule
\end{tabular}
\label{fig:q1_chart}
\end{table}

\subsection{Configuration}
We evaluate our framework across six LLMs, covering a diverse spectrum of sizes and categories. Specifically, we include: (1) \textit{Gemma3-4B} and (2) \textit{Gemma3-12B} as mid-sized open-source LLMs, (3) \textit{Mistral-Nemo-12B} as a domain-specific open-source model, (4) \textit{LLaMA3-70B} as a large-scale open-source model, (5) \textit{GPT-OSS-20B} as an open-source reasoning model, and (6) \textit{GPT-4o} as a representative closed-source LLM. For evaluation under the \emph{LLM-as-Judge} paradigm, we adopt \textit{Mixtral-8x7B}, a mixture-of-experts model with 57B parameters (7B active during inference), selected for its strong reasoning and comparative assessment capabilities. 

We assess personalization across three experimental settings: \textit{Direct Inference}, where the model generates responses without additional context; \textit{User Modeling}, where user-specific information is explicitly incorporated into the prompt; and \textit{Feature Selection}, where only task-relevant features are included. For memory updates, we investigate the effect of Direct Inference compared to configurations where the context includes both the Long-Term and Short-Term Memory. To further analyze the role of memory updates, we conduct an ablation study with two configurations: Single Inference, in which the user model is updated and a personalized response is generated within a single pass using structured output; and Two Inference, where these steps are performed sequentially in two separate passes. 

\begin{figure*}[t]
    \centering
    \begin{subfigure}[t]{0.24\textwidth}
        \centering
        \includegraphics[width=\textwidth]{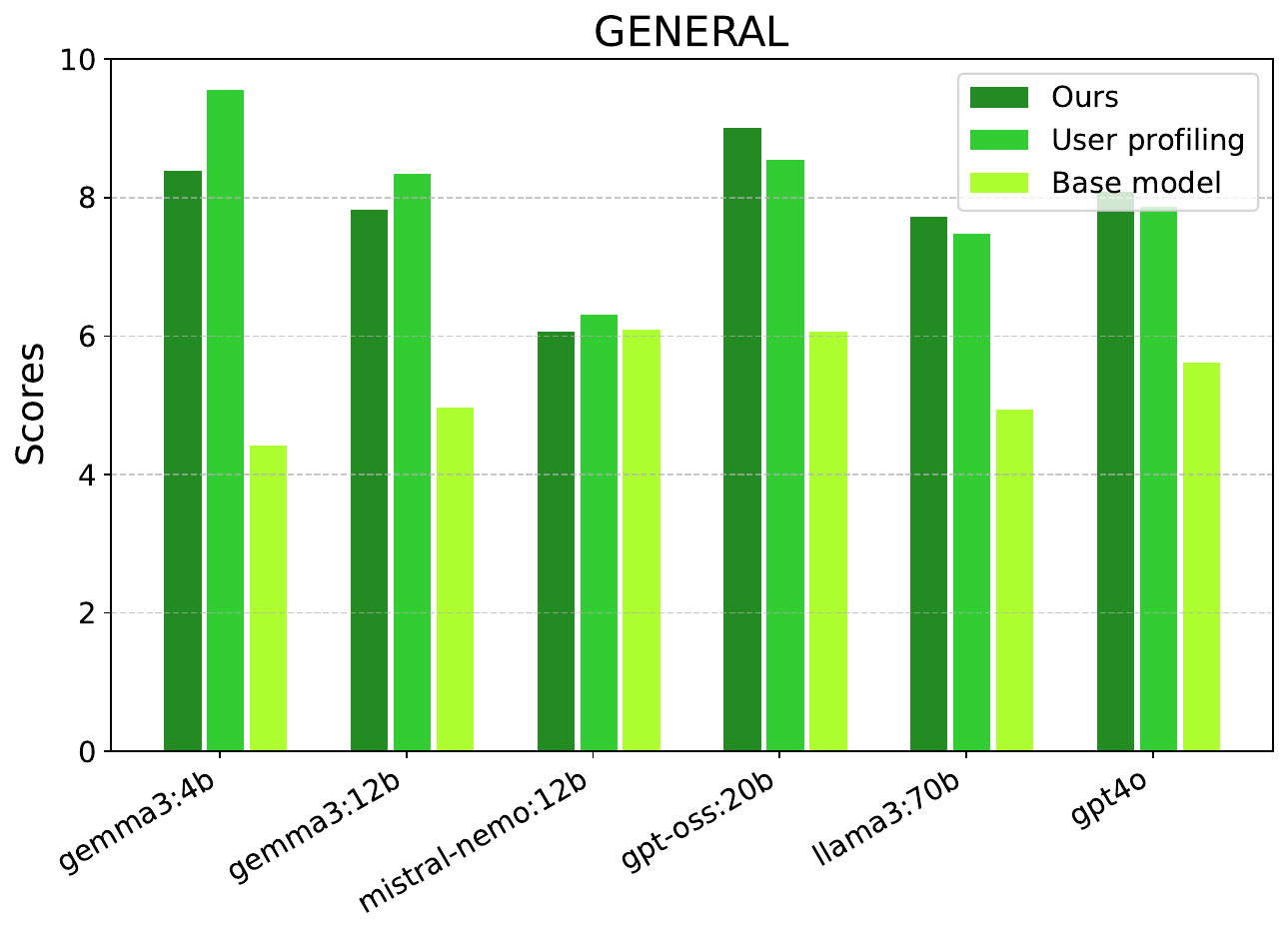}
        \caption{General replies (score).}
        \label{fig:persona-score-general}
    \end{subfigure}%
    ~ 
    \begin{subfigure}[t]{0.24\textwidth}
        \centering
        \includegraphics[width=\textwidth]{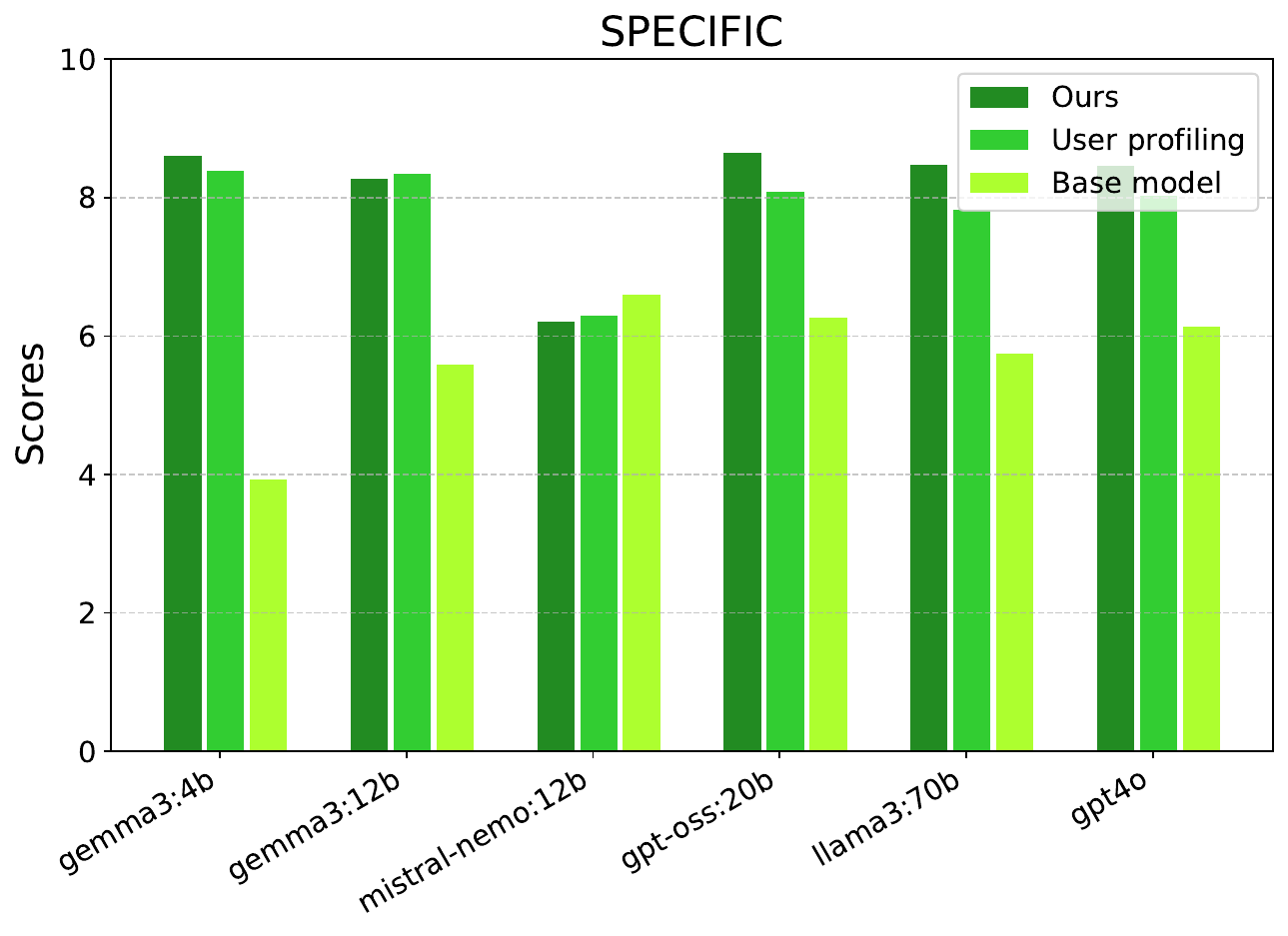}
        \caption{Specific replies (score).}
        \label{fig:persona-score-specific}
    \end{subfigure}%
    ~ 
    \begin{subfigure}[t]{0.24\textwidth}
        \centering
        \includegraphics[width=\textwidth]{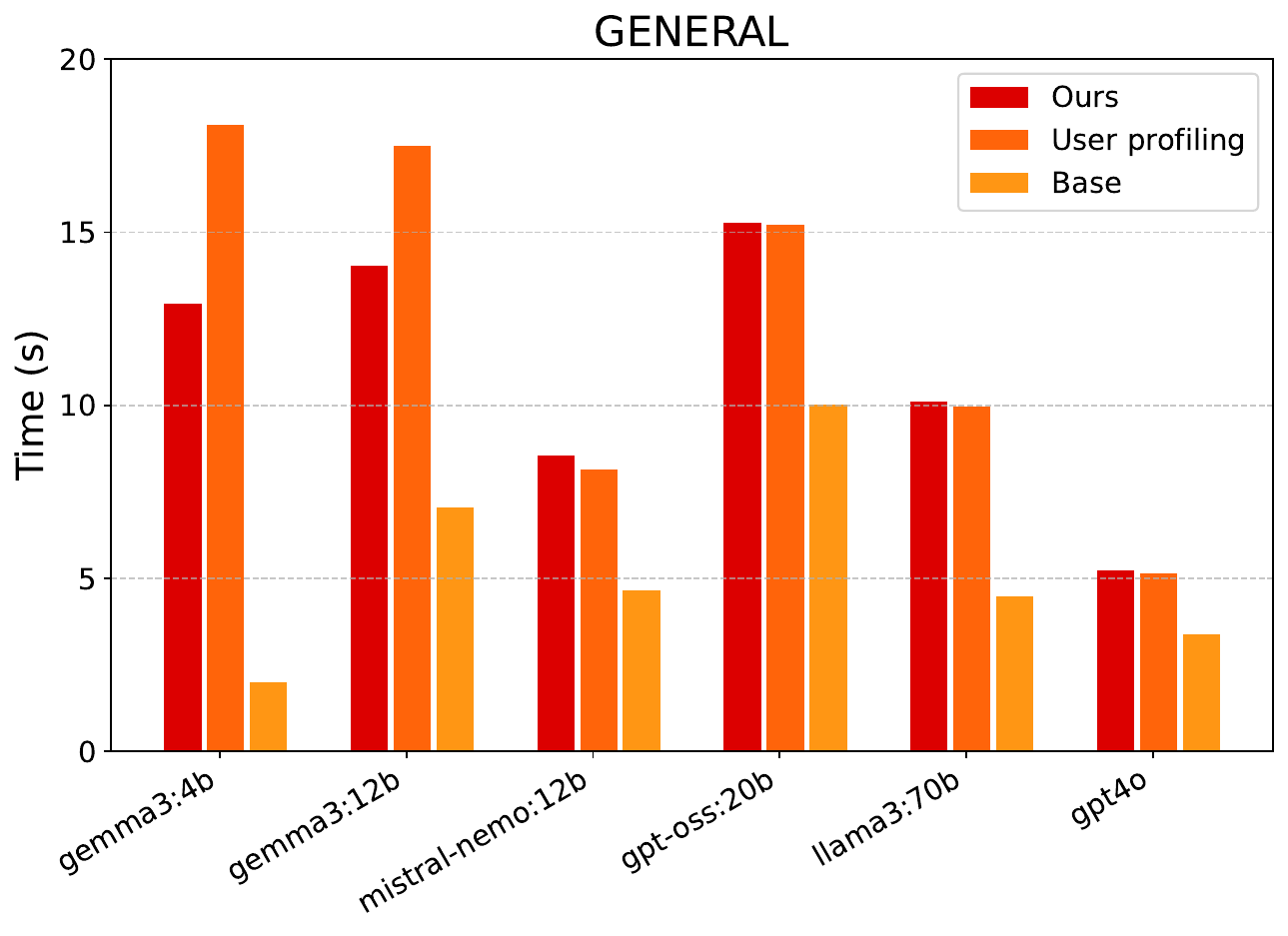}
        \caption{General replies (time).}
        \label{fig:persona-time-general}
    \end{subfigure}%
    ~ 
    \begin{subfigure}[t]{0.24\textwidth}
        \centering
        \includegraphics[width=\textwidth]{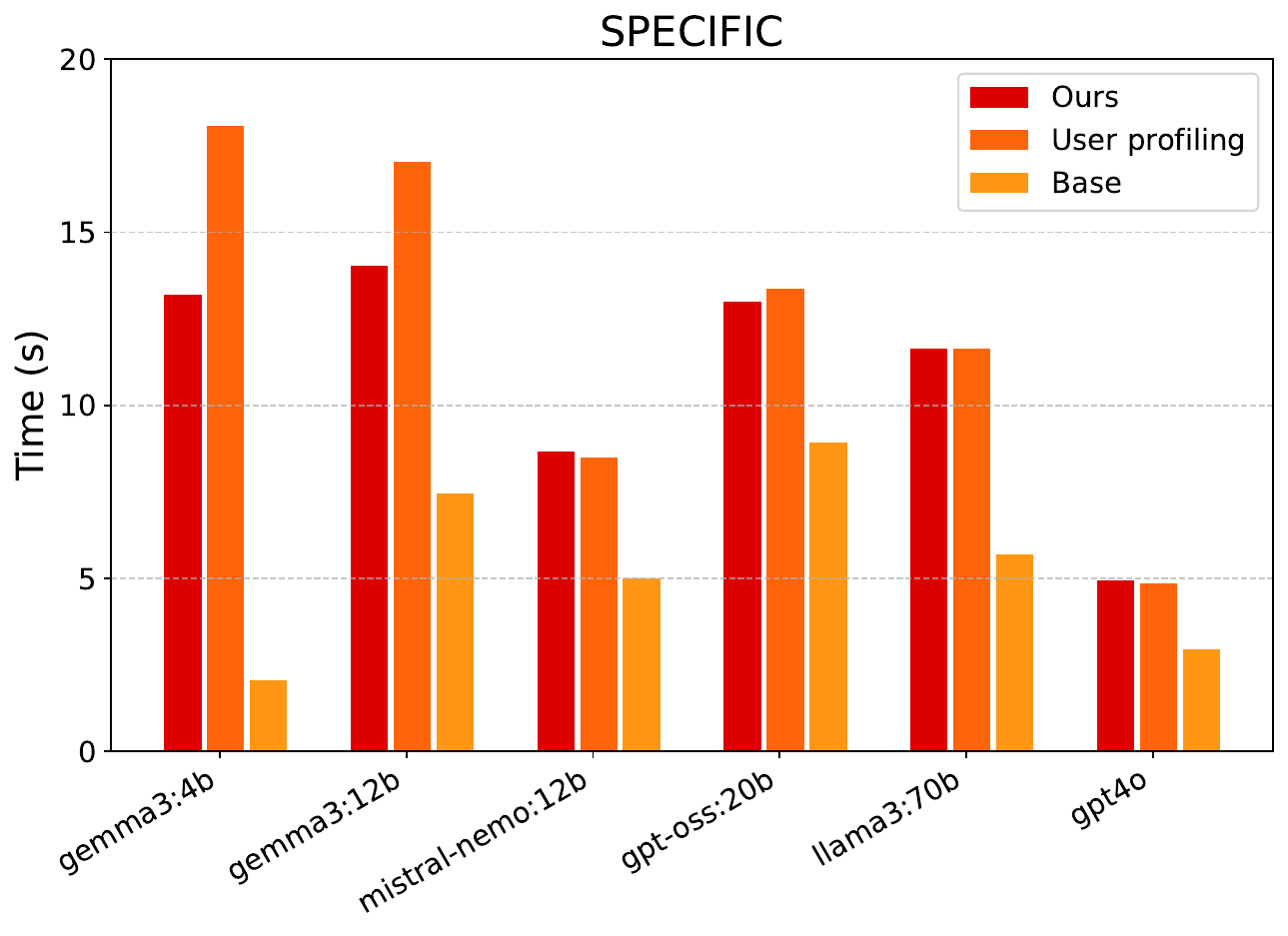}
        \caption{Specific replies (time).}
        \label{fig:persona-time-specific}
    \end{subfigure}
    \caption{Comparison of models on reply quality (\textcolor{red}{\textbf{Q3}}) and latency under different personalization settings (\textcolor{red}{\textbf{Q4}}) for PersonaFeedback\cite{personafeedback}. 
    \textit{Scores (a,b) are obtained using an LLM-as-a-Judge, evaluating both answer quality and personalization. Latency (c,d) measures response time. 
    We compare three configurations: direct inference with the \textbf{base model} (no profile), inference with the full \textbf{user profile}, and \textbf{our method}, which retrieves only pertinent profile features via similarity with the query.}}
    \label{fig:persona-overall}
\end{figure*}

Finally, human evaluation was conducted with 20 participants (13 women and 7 men), aged 65--91 years ($M = 78.3$, $SD = 7.02$). Each participant engaged in a 15-minute interaction scenario designed to reflect the increasing digitization of everyday services. Two experimental conditions were considered: in the first, participants provided specific information to the system to assess its memorization capabilities; in the second, information was pre-supplied to the system to evaluate its ability to return this information accurately. Both scenarios included one time-dependent feature and one general feature. Beyond these structured tasks, participants were encouraged to interact with the system freely on topics of their choice. Note that we have received ethical approval to conduct research involving human participants.

\begin{table*}[t]
\centering
\caption{\textbf{Ablation Study on Memory Update with GPT-4o over different configurations and different the inference settings (\textcolor{red}{\textbf{Q2}}).} The two-inference approach generally outperforms the single-inference variant, showing higher ROUGE scores, better session similarity, and improved effectiveness when combining short- and long-term memory. It also reduces latency by running inferences in parallel while maintaining task specialization and accuracy.}
\begin{tabular}{lccc|ccc|ccc|cc}
\toprule
\textbf{Configuration} & \multicolumn{3}{c}{\textbf{ROUGE-1}} & \multicolumn{3}{c}{\textbf{ROUGE-2}} & \multicolumn{3}{c}{\textbf{ROUGE-L}} & \multicolumn{2}{c}{\textbf{Session Similarity}}    \\  \cmidrule{2-4} \cmidrule{5-7} \cmidrule{8-10} \cmidrule{10-12}
                       & \textbf{P}       & \textbf{R}       & \textbf{F1}               & \textbf{P}        & \textbf{R}       & \textbf{F1}              & \textbf{P}       & \textbf{R}       & \textbf{F1}               & \textbf{Cosine} & \textbf{LaaJ} \\ \midrule \rowcolor{gray!15}
\multicolumn{12}{c}{Single Inference}                                                                                                     \\ \midrule \rowcolor{yellow!5}
Direct Inference              & 0.207   & 0.487   & 0.270            & 0.0890   & 0.214   & 0.119           & 0.198   & 0.467   & 0.258            & 0.854   &     6.81       \\ \rowcolor{yellow!15}
+ Long Term Memory     & 0.316   & 0.380   & 0.304            & 0.141    & 0.188   & 0.132           & 0.301   & 0.364   & 0.289            & 0.790  &      6.42       \\ \rowcolor{yellow!25}
+ Short Term Memory    & 0.212   & 0.379   & 0.250            & 0.081    & 0.164   & 0.100           & 0.200   & 0.359   & 0.237            & 0.864  &        5.67    \\ \midrule \rowcolor{gray!15}
\multicolumn{12}{c}{Two Inference}                                                                                                      \\ \midrule \rowcolor{green!5}
Direct Inference            & 0.458   & 0.232   & 0.272            & 0.225    & 0.096   & 0.118           & 0.429   & 0.216   & 0.253            & 0.833 &    \textbf{8.37}          \\ \rowcolor{green!20}
+ Long Term Memory     & 0.469   & 0.211   & 0.264            & 0.217    & 0.081   & 0.104           & 0.430   & 0.194   & 0.242            & 0.843  &      7.94       \\ \rowcolor{green!30}
+ Short Term Memory    & 0.428   & 0.337   & \textbf{0.344}   & 0.214    & 0.154   & \textbf{0.160}  & 0.396   & 0.314   & \textbf{0.318}   & \textbf{0.877}   & 8.06 \\ \bottomrule
\end{tabular}
\label{tab:rouge-results}
\end{table*}

\section{Results}

\subsection{Comparative Study}

\subsubsection*{\textcolor{blue}{\textbf{Perception module exhibits very high performance and very low latency across tasks}}}

The perception module processes the 45 videos streams received by the camera. Each video frame in the stream is passed through the face detection model and the result has been aggregated through majority and average norm of the prediction. As shown in \Cref{fig:q1_chart}, the module achieves close to 95\% accuracy for both face detection and 90\% accuracy for speaker recognition, while user retrieval reaches close to 98\% accuracy. Notably, the processing latency for each task is less or close to 3 ms, demonstrating the module's suitability for real-time applications. \Cref{fig:exmp} shows an example of speaker detection performed by the module, where the face grid highlights the detected speakers.

\medbreak

\subsubsection*{\textcolor{blue}{\textbf{The framework improves performance of personalization across all open-source and black-box LLMs}}}

In \Cref{fig:persona-score-general,fig:persona-score-specific}, we report the performance of our personalization method compared with the baseline across different open-source and black-box LLMs under two modes defined in the PersonaFeedback dataset: a general conversation setting and a more specific domain-oriented setting. Our results show that the proposed framework consistently outperforms the baselines in the specific mode across all evaluated LLMs. For the general mode, our framework achieves stronger improvements on larger LLMs, whereas for smaller LLMs, directly providing the entire user profile tends to yield better results. Furthermore, as illustrated in \Cref{fig:persona-time-general,fig:persona-score-specific}, we analyze latency and computational overhead. The proposed framework exhibits lower delay compared to the baseline but remains slower than direct inference, where fewer tokens are processed. This trade-off reflects the balance between reduced inference time and the accuracy gains achieved through personalization.

\medbreak

\subsubsection*{\textcolor{blue}{\textbf{Both short-term and long-term memory improve performance across LLMs at the expense of latency}}}

\begin{figure}[t]
    \centering
    \begin{subfigure}[t]{0.24\textwidth}
        \centering
        \includegraphics[width=\textwidth]{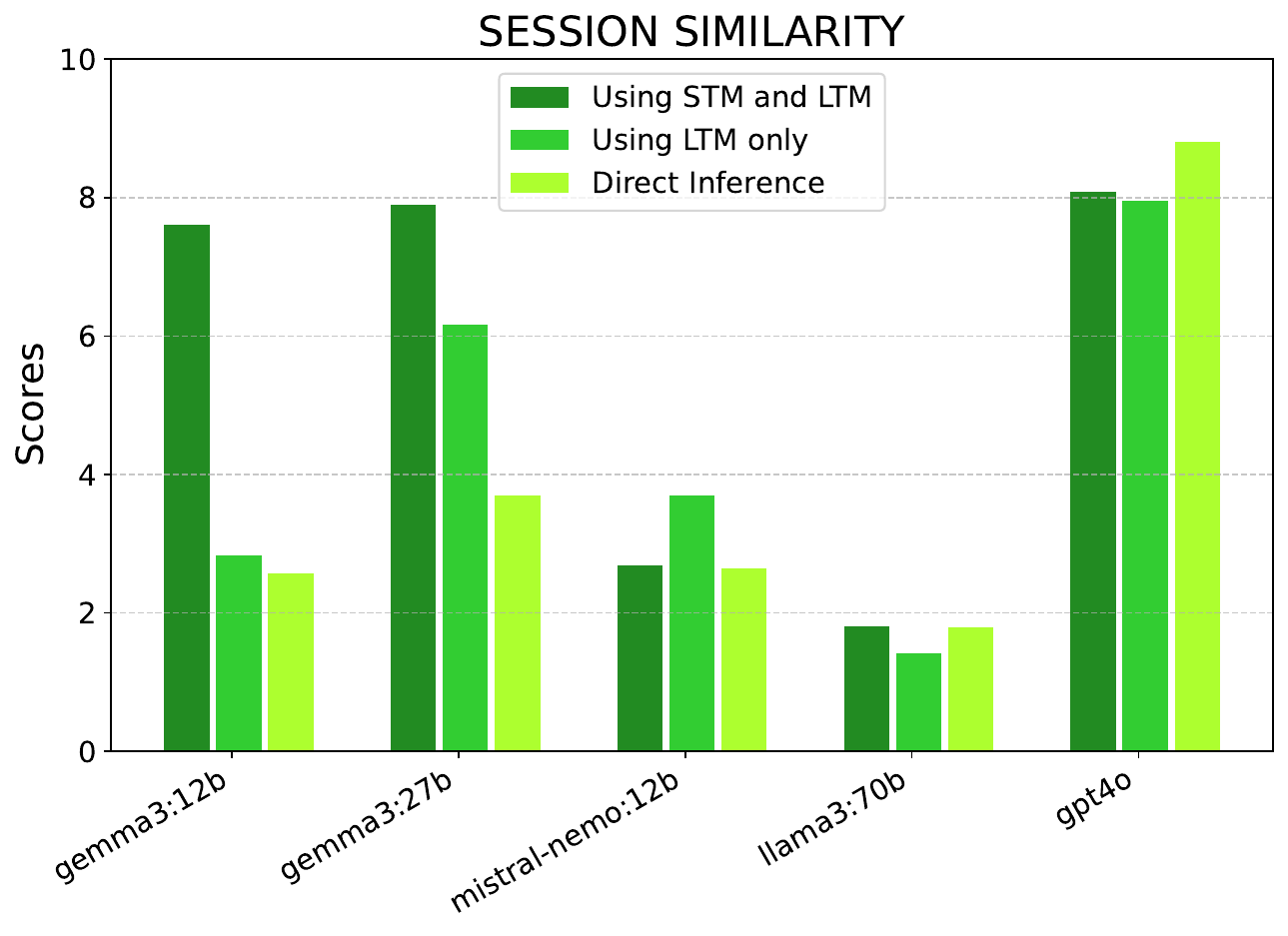}
        \caption{}
        \label{fig:locomo-session-sim}
    \end{subfigure}%
    ~ 
    \begin{subfigure}[t]{0.24\textwidth}
        \centering
        \includegraphics[width=\textwidth]{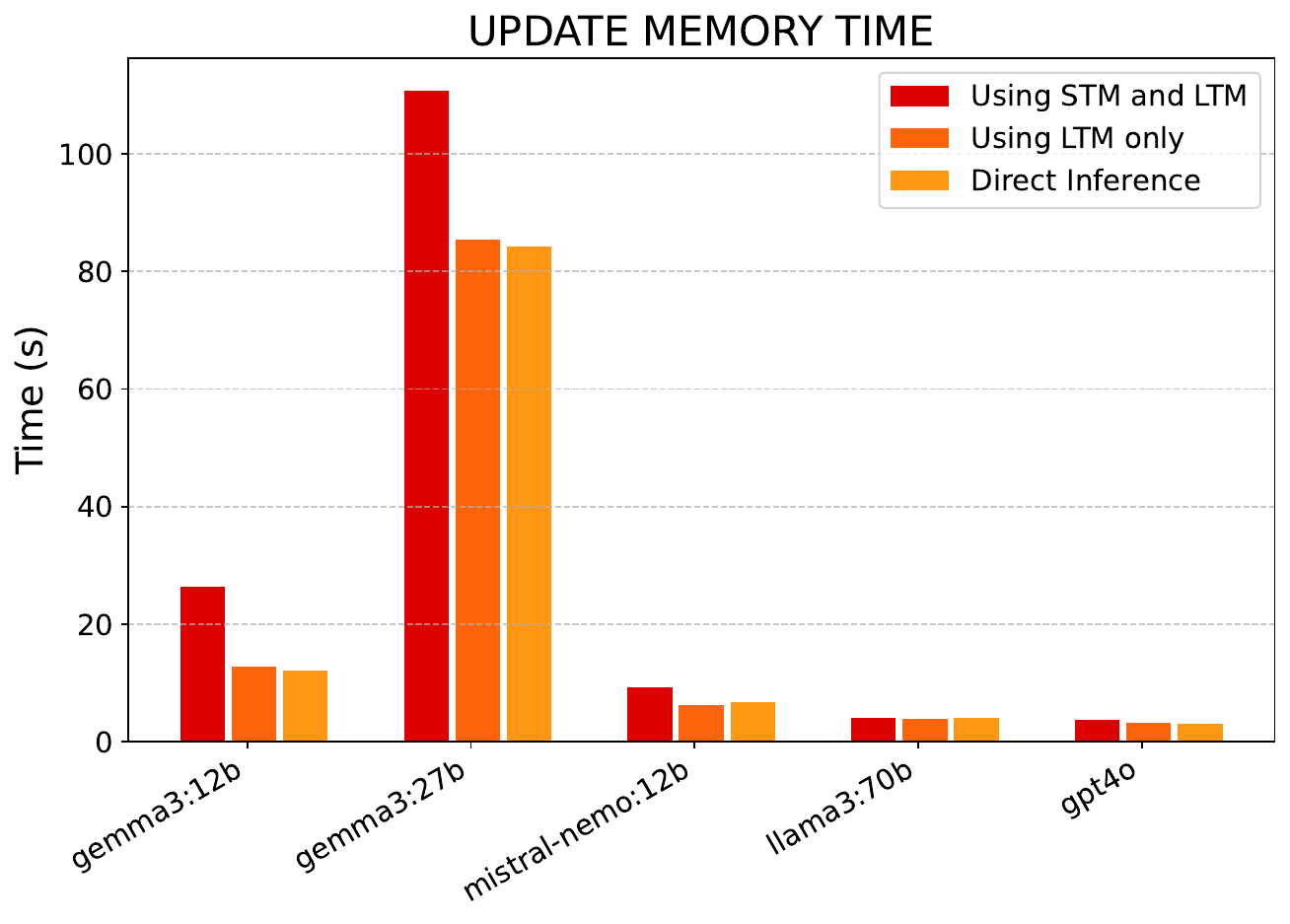}
        \caption{}
        \label{fig:locomo-time-update}
    \end{subfigure}
    \caption{\textbf{Evaluation of memory mechanisms in LoCoMo (\textcolor{red}{\textbf{Q2}}).} (a) Session similarity across different models under varying memory configurations. (b) Time required for memory updates across the same models and configurations (\textcolor{red}{\textbf{Q4}}).}
    \label{fig:locomo}
\end{figure}

As shown in \Cref{fig:locomo-session-sim}, incorporating both short-term and long-term memory into the system enhances feature extraction based on session-level semantic similarity for models such as Gemma and LLaMA. For Mistral, long-term memory proves more effective, while GPT-4 can reliably extract the necessary profiling information directly from the query without relying on external memory. However, as illustrated in \Cref{fig:locomo-time-update}, introducing additional tokens from short- and long-term memory inevitably increases latency, reflecting the trade-off between improved personalization and computational efficiency.

\subsection{Ablation Study}

\subsubsection*{\textcolor{blue}{\textbf{Structured output with a single inference does not outperform two parallel inferences}}}

As illustrated in \Cref{fig:interface}, our framework employs two LLM inferences: one generates the personalized response (\textit{Inf 2}), and the other updates the user profile (\textit{Inf 1}). With the advent of structured output—a paradigm that enforces LLMs to produce responses in a specified format—these tasks can be merged into a single inference. We evaluate this design by comparing GPT-4o under single- and two-inference settings. As shown in \Cref{tab:rouge-results}, the two-inference setting yields higher ROUGE F1 scores, except in the long-term scenario where the single-inference variant performs slightly better. This difference is more pronounced in LLM-as-a-Judge (LaaJ), with gains of up to three points. Across context configurations, combining short- and long-term memory proves most effective, improving both ROUGE and cosine similarity. In terms of session similarity, the two-inference approach consistently outperforms the single-inference variant. Finally, executing inferences separately but in parallel reduces latency while preserving task specialization, thereby achieving both lower delay and higher accuracy.

\medbreak

\subsubsection*{\textcolor{blue}{\textbf{Open-source LLMs rival black-box models in feature extraction and memory updating, albeit with higher failure rates}}}

As shown in \Cref{fig:locomo-models}, we conduct an ablation across models on the LoCoMo dataset, evaluating performance with LLM-as-a-Judge. We find that open-source LLMs such as \textsc{Gemma-3 12B} and \textsc{Gemma-27B} achieve performance comparable to the black-box \textsc{GPT-4o}. However, these models exhibit higher rates of task failure: from 18\% missed observations for \textsc{Gemma-3 12B} to nearly 80\% for \textsc{Mistral-Nemo 12B}. The \textbf{session score} measures how well a model reconstructs the profile after a full interaction session, while the \textbf{observation score} quantifies the similarity between features extracted at a given turn and the ground-truth observations at that turn. No score is reported when no features are extracted during an observation turn. We additionally report the percentage of \textbf{missed observations}, defined as the proportion of turns with an observation where the model failed to extract any features.

\begin{table}[t]
\centering
\footnotesize
\caption{\textbf{Comparison of memory updating performance across models (\textcolor{red}{\textbf{Q2}}).} \textit{Scores are evaluated using LLM-as-a-Judge. 
}}
\begin{tabular}{lccc}
\toprule
\textbf{Model} & \textbf{Session Sim} & \textbf{Observation Sim} & \textbf{Missed Obs (\%)} \\ \midrule \rowcolor{yellow!30}
gemma3:12b & 7.6  & 7.9  & 17    \\\rowcolor{yellow!30}
gemma3:27b & 8.3  & 8.41 & 24    \\ \rowcolor{red!30}
mistral-nemo:12b & 2.68 & 4.62 & 76.7  \\ \rowcolor{red!30}
llama3:70b & 1.81 & 1.45 & 90.53 \\  \rowcolor{green!30}
gpt4o      & 8.8  & 9.16 & 1.96  \\
\bottomrule
\end{tabular}
\label{fig:locomo-models}
\end{table}

\subsection{Qualitative Evaluation}
\subsubsection*{\textcolor{blue}{\textbf{The system is rated highly usable and well-validated by experts, with strengths in personalization}}} 
\begin{figure}
    \centering
    \includegraphics[width=\linewidth]{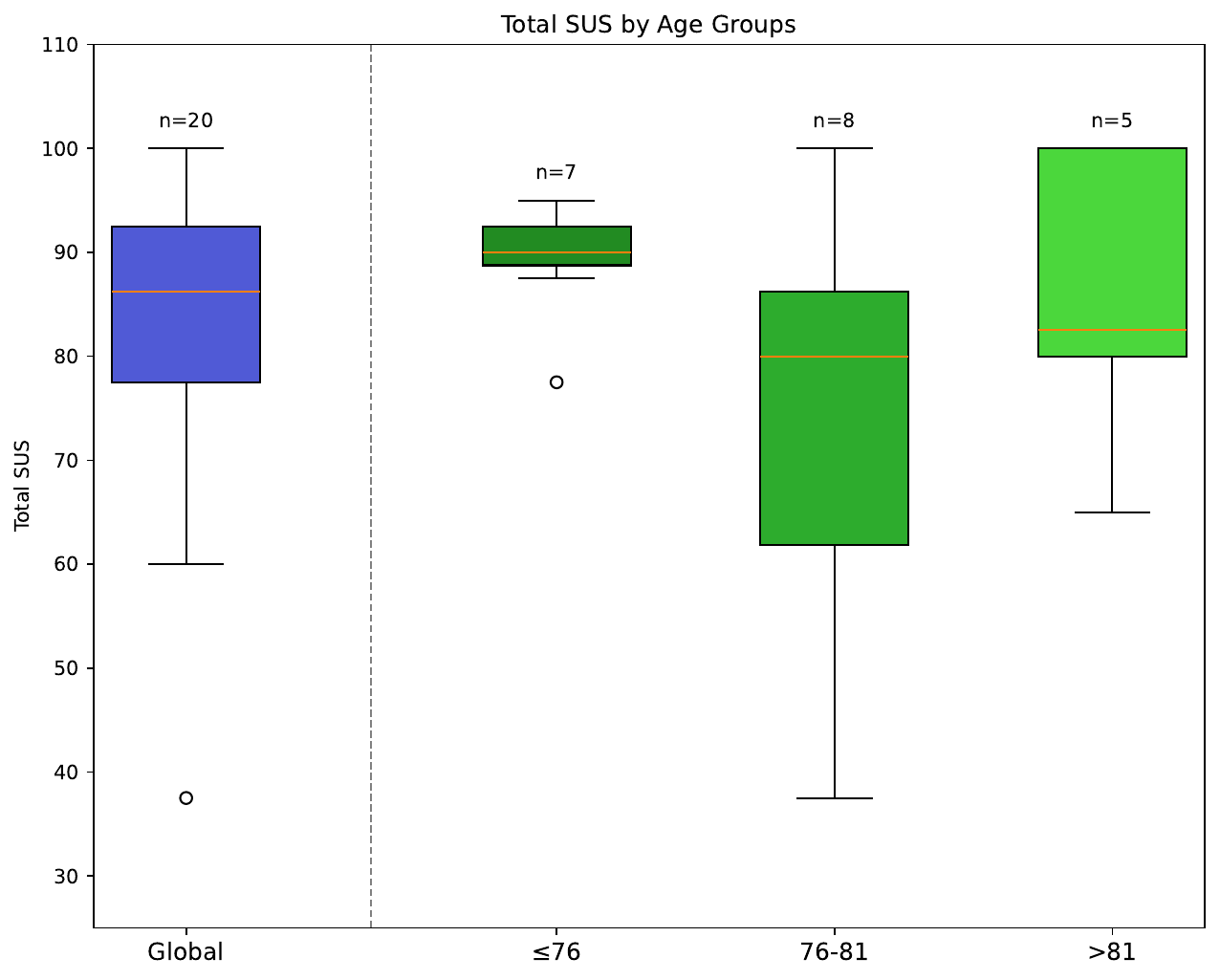}
    \caption{\textbf{Distribution of SUS score per age groups (\textcolor{red}{\textbf{Q5}}).} Participants reported high usability with a mean SUS score of 82.4 ($SD = 15.8$), placing the system in the ``Good’’ range and close to ``Excellent’’ while expert ratings averaged 4/5, confirming user feedback. }
    \label{fig:human_study}
\end{figure}

As shown in \Cref{fig:human_study}, the overall mean score across participants is 82.4 (SD = 15.8), which places the system in the category of acceptable usability (threshold: 72/100). According to \cite{bangor2008evaluationsus}, this corresponds to the “Good” range ($>$73) and approaches the “Excellent” range ($>$85). Expert evaluation also confirmed these results, with two experts rating the system 3.95 and 4.05 on a 5-point scale, yielding an average of 4/5.
With respect to information delivery and memorization, the system generally performed well, but showed weaknesses in recalling precise time-related information. For instance, it was unable to memorize exact appointment times, often replacing them with vague adverbs such as “soon.” To our surprise, users reported being highly satisfied with the personalization of the dialogue and the system’s ability to adapt in real time to both positive and negative feedback. However, participants also expressed frustration with certain limitations, particularly the system’s inability to access the internet and its restricted temporal knowledge (e.g., current weather, updated transportation fares). The tendency of the system to congratulate users on the quality of their questions was frequently mentioned—sometimes positively, sometimes negatively. Both users and experts also appreciated the system’s ability to answer “I don’t know” rather than inventing incorrect information.

\section{Implementation}

\begin{figure}
    \centering
    \includegraphics[width=\linewidth]{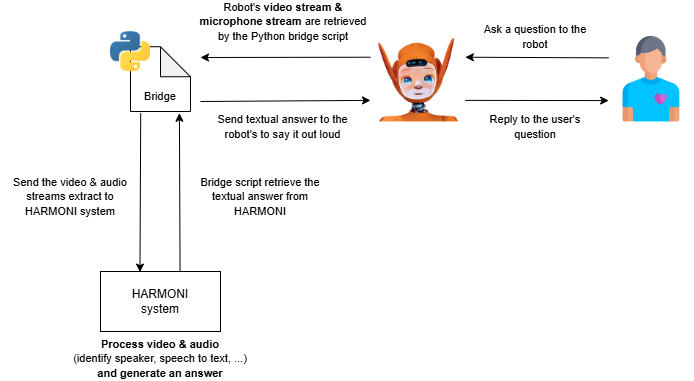}
    \caption{HARMONI x Miroka bridge architecture}
    \label{fig:imp}
\end{figure}

To demonstrate HARMONI’s applicability on a real robotic platform, we integrated it into Mirokai, a socially assistive robot from Enchanted Tools~\cite{enchantedtools-mirokai}, enabling multi-modal social interaction with an embodied agent. As shown in \Cref{fig:imp}, we designed an architecture in which HARMONI runs on a local machine and communicates with the Mirokai robot through a Python bridge script. To interact with and control the robot (using REST and WebSocket APIs), this script utilizes the pymirokai~\cite{pymirokai} library from Enchanted Tools. Audio and video captured by the robot are saved and sent to the API endpoints of HARMONI’s Video Interface app, which is started beforehand.
The first endpoint processes multimodal user input by transcribing speech, identifying the active speaker, and updating long-term user memory. The resulting transcription is then forwarded to another endpoint, which produces a context-aware response based on the current interaction and the user’s stored profile. Lastly, the generated textual response is sent to the Mirokai robot, which vocalizes it using its own text-to-speech system. This interaction loop enables the robot to engage in personalized, multi-turn conversations, illustrating how HARMONI’s capabilities transfer to an embodied robotic agent.
While this implementation validates the feasibility of deploying HARMONI on a social robot, some technical limitations remain. Audio and video streams currently rely on short recordings, which limits interaction fluidity and raises privacy concerns. Future work aims to enable real-time streaming, improve privacy by avoiding stored data, and activate perception modules only when user activity is detected.

\section{Conclusion}
We presented HARMONI, a multimodal personalization framework for multi-user human-robot interaction. By jointly leveraging perception, user modeling, world modeling, and generation, HARMONI enables robust reasoning and long-term personalization in dynamic settings. Experiments across multiple datasets and a real-world nursing home study demonstrate strong gains in speaker identification ($\sim$90\%), user profile retrieval ($\sim$98\%), and personalized response generation ($+$4/10), consistently surpassing baselines. Human participants rated the framework highly (82.4/100), and experts validated its effectiveness (4/5). Thanks to its modular design, HARMONI can manage uncertainty and adapt seamlessly to user needs, with broad potential in healthcare, education, and collaborative work. Future directions include extending to multi-agent interactions, incorporating richer affective and contextual signals, and studying the long-term impact of personalization on trust and engagement-paving the way for socially intelligent robotic systems.

\newpage

\bibliographystyle{unsrt}
\bibliography{references}










\end{document}